# SLCRF: Subspace Learning with Conditional Random Field for Hyperspectral Image Classification

Yun Cao, Jie Mei, Yuebin Wang, Liqiang Zhang, Junhuan Peng, Bing Zhang, Lihua Li, and Yibo Zheng

*Abstract*— Subspace learning (SL) plays an important role in hyperspectral image (HSI) classification, since it can provide an effective solution to reduce the redundant information in the image pixels of HSIs. Previous works about SL aim to improve the accuracy of HSI recognition. Using a large number of labeled samples, related methods can train the parameters of the proposed solutions to obtain better representations of HSI pixels. However, the data instances may not be sufficient enough to learn a precise model for HSI classification in real applications. Moreover, it is well-known that it takes much time, labor and human expertise to label HSI images. To avoid the aforementioned problems, a novel SL method that includes the probability assumption called subspace learning with conditional random field (SLCRF) is developed. In SLCRF, first, the 3D convolutional autoencoder (3DCAE) is introduced to remove the redundant information in HSI pixels. In addition, the relationships are also constructed using the spectral-spatial information among the adjacent pixels. Then, the conditional random field (CRF) framework can be constructed and further embedded into the HSI SL procedure with the semi-supervised approach. Through the linearized alternating direction method termed LADMAP, the objective function of SLCRF is optimized using a defined iterative algorithm. The proposed method is comprehensively evaluated using the challenging public HSI datasets. We can achieve state-of-the-art performance using these HSI sets.

*Index Terms*—Hyperspectral image classification, subspace learning, relationship construction, conditional random field, optimization.

## I. INTRODUCTION

Hyperspectral image classification, as a hot and significant research topic, has received extensive attention in the fields of remote sensing and computer vision [1], [2]. The difficulty of estimating the relationship between two pixels is the prominent challenge in HSI recognition due to its high dimensionality [3]-[8]. Hence, SL plays an important role in HSI classification, since it is an effective solution to reduce the redundant information of HSI pixels [9].

The related works about SL have appeared in the fields of computer vision, as well as in remote sensing. In computer vision, there are a variety of SL models and approaches to transform the high dimensional image data into more compact representations [10-11]. Dimensionality reduction maps the high- dimensional data of the original feature to a low-dimensional subspace, which can be equal to SL. Some classical approaches have been developed to effectively preserve the statistical properties, including principal component analysis (PCA) [12] and independent component analysis [13]. Manifold learning is usually adopted and embedded into the SL procedure [14]-[16]. To discover the nonlinear degrees of freedom that may underlie the complex natural observations, ISOMAP was developed in [14]. With respect to the Laplacian eigenmap (LE), a geometrically motivated algorithm was proposed to provide a computationally efficient approach for SL [15]. Locally linear embedding was designed to find compact expressions of high-dimensional features using the unsupervised learning algorithm [16]. Moreover, a feature representation learning method was adopted by combining image understanding, feature correlation and feature learning to learn the underlying subspace [10].

In the related field of computer vision, many SL approaches have been introduced to perform related tasks, including facial recognition, image understanding, and image segmentation. However, directly applying these methods to remote sensing may degrade the accuracy of remote sensing image recognition. Most importantly, the structures of the data sources are different. HSIs have distinctive data components and image features. Hence, a specially designed SL algorithm should be constructed to better recognize HSIs. A dimensionality reduction method with an enhanced alignment technique was proposed to maximize the class distances and retain the inherent geometric structures of the features using labeled and unlabeled training samples [9]. For the class separability, neighborhood data structures and nearby feature line measurements were both introduced to determine the transformation of the dimensionality reduction in the eigenspace. To take advantage of the spectral information and the spatial correlation between HSI pixels and further avoid the "salt and pepper" problem, researchers developed the PSASL method [17]. A similar method also appeared in the work of [18] in which a projection learning

Y. Cao, Y. Wang, J. Peng, and L. Li are with the School of Land Science and Technology, China University of Geosciences, Beijing 100083, China. (cy12160019@163.com, xxgcdxwyb@163.com, pengjunhuan@163.com, lihuali@cugb.edu.cn)

J. Mei, L. Zhang and Y. Zheng are with the Faculty of Geographical Science, State Key Laboratory of Remote Sensing Science, Beijing Normal University, Beijing 100875, China (e-mail: meijie@mail.bnu.edu.cn; zhanglq@bnu.edu.cn; bnuzyb@163.com ).

B. Zhang is with the Institute of Remote Sensing and Digital Earth, Chinese Academy of Sciences, Beijing 100094, China (e-mail: zb@ceode.ac.cn).

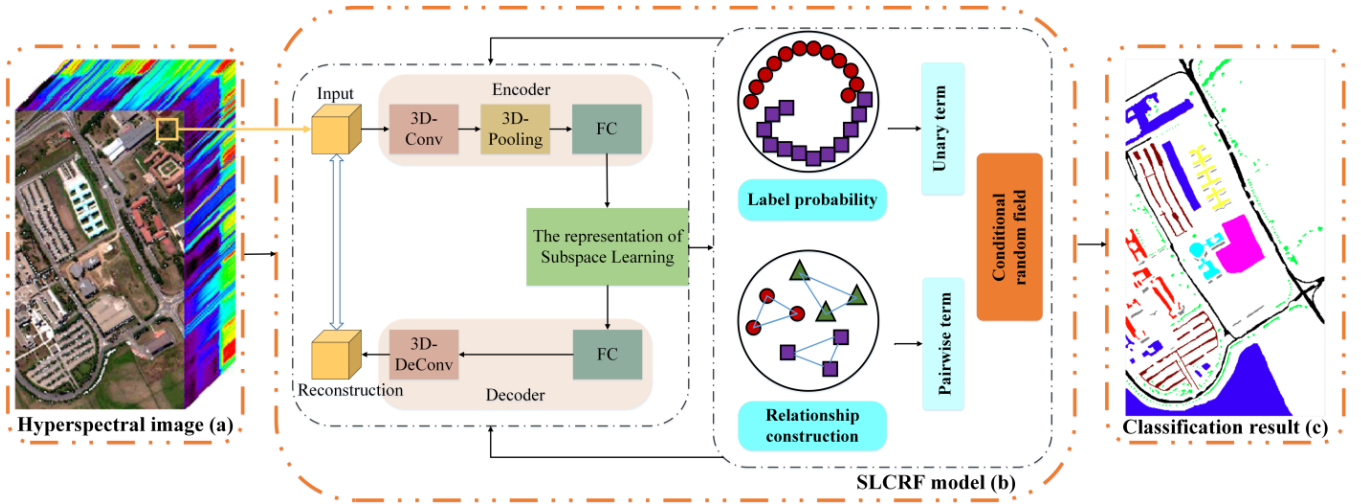

Fig. 1. The SLCRF workflow for HSI classification. (a) The HSI. (b) The learning procedure of the SLCRF. (c) The classification results. In (b), the representation of subspace learning is obtained from the 3DCAE. Then, the conditional random field framework is constructed and further embedded into the SL procedure.

approach with both local and global consistency constraints was proposed for aerial image recognition.

Previous works about SL aim at improving the accuracy of HSI recognition. With a great deal of labeled training samples, some methods can train the parameters of the proposed solutions to obtain better representations of HSI pixels. Convolutional neural networks (CNNs) can be used for HSI classification by combining with the softmax classifier and cross-entropy loss in a supervised way. Hu *et al.* [19] employed deep CNNs using a softmax classifier for HSI classification in a pixel-level. Lee *et al.* [20] proposed a wider and deeper CNN that has a multi-scale convolutional filter to obtain initial spatial and spectral feature maps. The maps are then fed through a fully convolutional network that produces the labels of HSI with a softmax classifier. Such an end-to-end classification system can classify the HSI in a supervised way. However, supervised methods need large training samples. In real applications, the data instances may not be sufficient enough to learn a precise model for HSI recognition. Moreover, it is well known that it takes much time, labor and human expertise to label HSI images [21].

Recent research on self-supervised learning [22-24] has shown us that data information can be considered as a self-supervision source for SL. In this way, better features of HSI pixels can be obtained without large amounts of manual labeling. The information of HSI data mainly includes the probabilities of pixels belonging to different categories and the relationships between pixels. To better embed the above data information into our proposed method, a probability model can be introduced to obtain better representations of HSI pixels. CRF is a good way that can incorporate different image features for classification [25], [26]. In the work of [26], a saliency method was developed that jointly learns a CRF and a visual dictionary. The convolutional CRF using the confidence map was proposed to achieve better focus region detection [27]. The advantage of the CRF was further extended by using convolutional neural networks for low-level feature extraction [28].

SLCRF is developed to obtain the subspace of the HSI pixels using the semi-supervised approach in this paper. To increase the accuracy of the SL of HSIs, the CRF is embedded into the SLCRF learning framework. First, 3DCAE is trained to remove the redundant information of HSI pixels through the reconstruction error. Thus, the latent subspace can be obtained from the original data. To obtain accurate subspace representation, the relationship matrix is further formulated by combining the relationships of the latent space and the spatial distance. However, good representation of the latent subspace and good relationships among pixels cannot be obtained without the probability maximization. The CRF framework can provide a good approach to enhance the performance of SL. The CRF unary and pairwise terms are constructed using the probabilities of pixels belonging to different categories and the relationships between pixels. Then, a novel approach termed SLCRF is proposed to incorporate the CRF framework into the SL procedure. The SL, the relationship construction and the semi-supervised constraints that are included in the CRF can build a unified objective function. The objective function of SLCRF can further use an iterative optimization algorithm. The developed method is tested using three public adopted HSI datasets. The results can validate that the SLCRF can perform better than the related HSI classification approaches. The workflow of the SLCRF is shown in Fig. 1.

The main contributions of the SLCRF can be further summarized in the following.

1) With the learned subspace, the probabilities of pixels belonging to different categories and the relationships between the neighboring pixels are constructed to form the corresponding terms of the CRF. With the CRF, the better HSI subspace can be obtained to alleviate the need for labeled samples.

2) SLCRF integrates the SL, the relationship construction and the semi-supervised constraints that are defined in the CRF to form the objective function. An iterative optimization algorithm is used to solve the objective function of the SLCRF.

3) SLCRF is an efficient end-to-end trained framework. 3DCAE allows for SL. The CRF is further embedded into SL to provide feedback information for better SL. It can outperform the state-of-the-art approaches using three image datasets.

For clarity, the notations and definitions of this paper are illustrated in Table I.

Table I
NOTATIONS AND DEFINITIONS OF SLCRF

| Notation | Definition |
|---|---|
| $\mathbf{X}$ | Data matrix. $\mathbf{X} \in \mathbb{R}^{b \times b \times d}$ |
| $\mathbf{X}^{(M)}$ | The reconstructed matrix of $\mathbf{X}$. $\mathbf{X}^{(M)} \in \mathbb{R}^{b \times b \times d}$ |
| $x^{(M_1)}$ | The representation in the latent subspace of $\mathbf{X}$. |
| $\mathbf{S}$ | The matrix of the relationship between pixels. $\mathbf{S} \in \mathbb{R}^{n \times n}$ |
| $\mathbf{Z}$ | The matrix of the sparse representation. $\mathbf{Z} \in \mathbb{R}^{n \times n}$ |
| $\theta$ | The parameters of CAE. |
| $\mathbf{Y}$ | The predicted labels of pixels. $\mathbf{Y} \in \mathbb{R}^{c \times n}$ |
| $\mathbf{h}_1$ | The parameters of the unary term of the CRF. |
| $\mathbf{h}_2$ | The scalar that measures the weight of the labeling smoothness. |
| $\mathcal{G}$ | The adaptive graph. |
| $\alpha, \beta, \gamma, \eta$ | Balance the corresponding terms. |
| $\lambda_1, \lambda_2$ | Balance the corresponding terms. |
| $d$ | The number of bands. |
| $n$ | The number of samples. |
| $P_i$ | The position of pixel $i$. |

## II. SUBSPACE LEARNING WITH THE CONDITIONAL RANDOM FIELD (SLCRF)

Given an HSI, the high-dimension HSI dataset can be constructed in a mathematical format. Let $P_{(x,y)} \in \mathbb{R}^{d \times 1}$ be a pixel located at $(x, y)$ of the HSI, where $d$ is the number of spectrum bands. The spatial neighborhood of $P_{(x,y)}$ is defined as $\mathbf{X} \in \mathbb{R}^{b \times b \times d}$, where $b \times b$ is the spatial neighborhood of pixel $(x, y)$.

### A. Definition of Subspace Learning

For an HSI, we use the 3DCAE framework to obtain its representation in the latent subspace, which consists of an encoder and a decoder. The encoder maps the original data $\mathbf{X} \in \mathbb{R}^{b \times b \times d}$ to the representation of latent subspace $x^{(M_1)} \in \mathbb{R}^{K \times 1}$, where $K$ represents the dimensions of the latent subspace. The decoder reconstructs an approximation $\mathbf{X}^{(M)} \in \mathbb{R}^{b \times b \times d}$ of $\mathbf{X}$ from $x^{(M_1)}$. There have $M$ layers in the 3DCAE, where the encoder has $M_1$ layers and decoder has $(M-M_1)$ layers. The encoder consists of 3D convolutional layers, 3D max-pooling layers, and fully connected layers to learn the representations of latent subspace. The decoder consists of 3D deconvolution layers and fully connected layers to reconstruct the original data. More details of 3DCAE can be referred to the Appendixes.

The objective function of 3DCAE can be formulated as follows:

$$\Theta_1(\theta) = \frac{1}{b \times b \times d} \sum_{x=0}^{b-1} \sum_{y=0}^{b-1} \sum_{z=0}^{d-1} \left| X_{x,y,z} - X_{x,y,z}^{(M)} \right|^2 + \frac{\alpha}{2} \|\mathbf{W}\|_F^2 \quad (1)$$

where $\alpha$ is the balancing term between the reconstruction error and the weight decay term. $\theta$ represents the set of all parameters in 3DCAE, and $\mathbf{W}$ consists of the weights in all layers. $X_{x,y,z}$ represents the value at position $(x, y, z)$ of the input $\mathbf{X} \in \mathbb{R}^{b \times b \times d}$, while $X_{x,y,z}^{(M)}$ represents the reconstructed value at position $(x, y, z)$. The weight decay term helps to decrease the magnitude of the weights and prevent over-fitting. Based on the 3DCAE [29], [30], equation (1) acquires the latent subspace from the original data $\mathbf{X}$ by training the 3DCAE through the reconstruction error. By minimizing the equation, we can acquire the subspace of the corresponding HSI.

### B. Relationships' Construction with Subspace Learning

To obtain accurate subspace representation, the relationship matrix $\mathbf{S}$ between pixels is constructed to enhance the SL, which is formulated by combining the relationships of the latent space and the spatial distance. The relationships of the latent space are constructed by using the sparse coding to obtain the sparse matrix, which describes the relationships in the latent subspace. Let $\mathbf{X}^{(M_1)} = [x_1^{(M_1)}, x_2^{(M_1)}, ..., x_n^{(M_1)}] \in \mathbb{R}^{K \times n}$ be the representation of latent subspace for all HSI pixels, where $n$ is the number of pixels. Considering $\mathbf{X}^{(M_1)}$ as the dictionary, the following equation is utilized to implement the sparse coding of latent subspace:

$$\Theta_2(\theta, \mathbf{Z}) = \left\| \mathbf{X}^{(M_1)} - \mathbf{X}^{(M_1)} \mathbf{Z} \right\|_F^2 + \beta \|\mathbf{Z}\|_1 \quad (2)$$

where $\beta$ is the balancing constant, $\| \ \|_F^2$ represents the Frobenius norm of a matrix, and $\| \ \|_1$ represents the L1 norm of a matrix. $\mathbf{Z}$ describes the relationships the latent subspace, and each column vector of $\mathbf{Z}$ is the sparse representation. To better estimate $\mathbf{Z}$, the initialization of $\mathbf{Z}$ is obtained by optimizing the equation $\left\| \mathbf{X}^{(M_1-1)} - \mathbf{X}^{(M_1-1)} \mathbf{Z} \right\|_F^2 + \beta \|\mathbf{Z}\|_1$ with $\mathbf{X}^{(M_1-1)}$ fixed. The relationship matrix $\mathbf{S}_1$ of the latent space can be formed as follows:

$$\mathbf{S}_1 = \left| \mathbf{Z} + \mathbf{Z}^T \right| / 2 \quad (3)$$

Second, the relationship matrix $\mathbf{S}_2$ of spatial distance is constructed to describe the relationships of the spatial distance:

$$\mathbf{S}_{2ij} = \begin{cases} \exp\left( -\frac{\|P_i - P_j\|_2^2}{\omega} \right), & x_i^{(M_1)} \in \mathcal{N}_k\left(x_j^{(M_1)}\right) \text{ or } x_j^{(M_1)} \in \mathcal{N}_k\left(x_i^{(M_1)}\right) \\ 0 & \text{otherwise} \end{cases} \quad (4)$$

where $\mathcal{N}_k\left(x_j^{(M_1)}\right)$ is the $k$-nearest neighbors of $x_i^{(M_1)}$ that are constructed using the spatial distance $\|P_i - P_j\|_2$. $\omega$ is an alternative parameter.

By combining the relationships of the latent space and the spatial distance, we can formulate the relationship matrix $\mathbf{S}$ to denote the relationships in subspace:

$$\mathbf{S} = \mathbf{S}_1 + \gamma \mathbf{S}_2 \quad (5)$$

where $\gamma$ is a constant.

### C. CRF Introduction for Subspace Learning

From the definition of the SL, we can obtain the representation in the latent subspace. However, good representation of the latent subspace and good relationships among pixels cannot be obtained without the probability maximization. The CRF can provide a better approach to enhance the performance of SL.

The CRF usually needs a graph to construct its terms including the unary term and binary term. In our method, the graph $\mathcal{G} = <\mathcal{V}, \mathcal{E}>$ is built to reflect the relationships among different HSI pixels. $\mathcal{V}$ refers to the representations of the latent subspace in an HSI. Each vertex corresponds to one representation of the latent subspace. $\mathcal{E}$ is the edges that are linked by the neighboring pixels. Its weights can be constructed using the matrix $\mathbf{S}$ that is defined in (5).

In a CRF, the unary term describes the possibility of one belonging to one certain category, and the pairwise term refers to the relationships between different pixels. The CRF ($\mathbf{X}^{(M_1)}$, $\mathbf{Y}$) can be formulated by a Gibbs distribution:

$$p\left(\mathbf{Y} \mid \mathbf{X}^{(M_1)}; \theta, \mathbf{Z}, \mathbf{h}\right) = \frac{1}{C} e^{-\sum_{x_i^{(M_1)} \in \mathcal{V}} E\left(x_i^{(M_1)}, \theta, \mathbf{Z}, \mathbf{h}\right)} \quad (6)$$

where $C$ is the partition function, and $E(\theta, \mathbf{Z}, \mathbf{h})$ is the energy function with the weight $\mathbf{h}$ and parameters $\theta$ and $\mathbf{Z}$. To optimize the model, it needs to maximize conditional distribution $p(\mathbf{Y} \mid \mathbf{X}^{(M_1)}; \theta, \mathbf{Z}, \mathbf{h})$:

$$\max_{\theta, \mathbf{Z}, \mathbf{h}} p\left(\mathbf{Y} \mid \mathbf{X}^{(M_1)}; \theta, \mathbf{Z}, \mathbf{h}\right) = \max_{\theta, \mathbf{Z}, \mathbf{h}} \frac{1}{C} e^{-\sum_{x_i^{(M_1)} \in \mathcal{V}} E\left(x_i^{(M_1)}, \theta, \mathbf{Z}, \mathbf{h}\right)} \quad (7)$$

which is equivalent to

$$\min_{\theta, \mathbf{Z}, \mathbf{h}} \sum_{x_i^{(M_1)} \in \mathcal{V}} E\left(x_i^{(M_1)}, \theta, \mathbf{Z}, \mathbf{h}\right) \quad (8)$$

Then, we can formulate a novel CRF model

$$E\left(x_i^{(M_1)}, \theta, \mathbf{Z}, \mathbf{h}\right) = \phi(\theta, \mathbf{h}_1) + \eta \varphi(\mathbf{Z}, \mathbf{h}_2) \quad (9)$$

where $\eta$ is the balancing constant, and $\mathbf{h} = \{\mathbf{h}_1, \mathbf{h}_2\}$ is the weight matrix. $\mathbf{h}_1$ is adopted to compute the categorical possibilities using the softmax function, and $\mathbf{h}_2$ measures the weight of the pairwise term. The functions $\phi$ is the unary term in the CRF:

$$\phi\left(\mathbf{X}^{(M_1)}, \mathbf{Y}, \theta, \mathbf{h}_1\right) = \sum_{x_i^{(M_1)} \in \mathcal{V}} -I(t) \log\left(p(\mathbf{y}_i / x_i^{(M_1)})\right)$$

where $\mathbf{h}_1 = \{\mathbf{W}^{\mathbf{h}1}, \mathbf{b}^{\mathbf{h}1}\}$. The value of $I(t)$ is 1 if $t$ equals to the desired label of pixel $i$; otherwise, its value is 0. $p(\mathbf{y}_i / x_i^{(M_1)})$ is achieved by using the softmax function $\Phi$:

$$p(\mathbf{y}_i / x_i^{(M_1)}) = \Phi\left(\mathbf{W}_t^{\mathbf{h}1} x_i^{(M_1)} + \mathbf{b}_t^{\mathbf{h}1}\right)$$

The functions $\varphi$ is the pairwise term in the CRF:

$$\varphi(\mathbf{Y}, \mathbf{Z}, \mathbf{h}_2) = \sum_{x_i^{(M_1)} \in \mathcal{V}} \sum_{x_j^{(M_1)} \in \mathcal{N}_k(x_i^{(M_1)})} -\mathbf{h}_2 \mathbf{S}_{i,j}, \quad \mathbf{h}_2 = \frac{(\mathbf{y}_i - \mathbf{y}_j)^2}{\sum_{x_j^{(M_1)} \in \mathcal{N}_k(x_i^{(M_1)})} \mathbf{S}_{i,j}}$$

where $\mathbf{h}_2$ is scalar measuring the weight of labeling smoothness.

### D. Objective Function Formulation of SLCRF

Given the training samples $\mathbf{X}$ and the corresponding predicated labels, the SLCRF method is to learn the SL parameters $\theta$ and the weights $\mathbf{h}$ in the CRF to maximize the joint likelihood of training samples. Here, we combine all the terms together and establish the SLCRF model:

$$\min_{\theta, \mathbf{Z}, \mathbf{h}} \Theta_1(\theta) + \lambda_1 \Theta_2(\theta, \mathbf{Z}) + \lambda_2 E(\theta, \mathbf{Z}, \mathbf{h})$$

where $\lambda_1 > 0, \lambda_2 > 0$ are the trade-off parameters. Equivalently, the SLCRF model can be defined using the following equation:

$$\min_{\theta, \mathbf{Z}, \mathbf{h}} \frac{1}{b \times b \times d} \sum_{x=0}^{b-1} \sum_{y=0}^{b-1} \sum_{z=0}^{d-1} \left(X_{x,y,z} - X_{x,y,z}^{(M)}\right)^2 + \frac{\alpha}{2} \|\mathbf{W}\|_F^2$$
$$+ \lambda_1 \left\|\mathbf{X}^{(M_1)} - \mathbf{X}^{(M_1)} \mathbf{Z}\right\|_F^2 + \lambda_1 \beta \|\mathbf{Z}\|_1$$
$$+ \lambda_2 \sum_{x_i^{(M_1)} \in \mathcal{V}} -I(t) \log\left(p(\mathbf{y}_i / x_i^{(M_1)})\right) + \lambda_2 \eta \sum_{x_i^{(M_1)} \in \mathcal{V}} \sum_{x_j^{(M_1)} \in \mathcal{N}_k(x_i^{(M_1)})} -\mathbf{h}_2 \mathbf{S}_{i,j}$$
(10)

Due to the close dependence of $\mathbf{S}$ on $\mathbf{Z}$, the auxiliary matrix $\mathbf{M} \in \mathbb{R}^{n \times n}$ is introduced to separate (10). Then, the whole objective function can be transferred into

$$\min_{\theta, \mathbf{Z}, \mathbf{h}} \frac{1}{b \times b \times d} \sum_{x=0}^{b-1} \sum_{y=0}^{b-1} \sum_{z=0}^{d-1} \left(X_{x,y,z} - X_{x,y,z}^{(M)}\right)^2 + \frac{\alpha}{2} \|\mathbf{W}\|_F^2$$
$$+ \lambda_1 \left\|\mathbf{X}^{(M_1)} - \mathbf{X}^{(M_1)} \mathbf{M}\right\|_F^2 + \lambda_1 \beta \|\mathbf{M}\|_1$$
$$+ \lambda_2 \sum_{x_i^{(M_1)} \in \mathcal{V}} -I(t) \log\left(p(\mathbf{y}_i / x_i^{(M_1)})\right) + \lambda_2 \eta \sum_{x_i^{(M_1)} \in \mathcal{V}} \sum_{x_j^{(M_1)} \in \mathcal{N}_k(x_i^{(M_1)})} -\mathbf{h}_2 \mathbf{S}_{i,j}$$
$$\text{s.t. } \mathbf{Z} = \mathbf{M}$$
(11)

### III. OPTIMIZATION OF SLCRF

To optimize the SLCRF model, the linearized alternating direction method termed LADMAP [31] is adopted. The corresponding augmented Lagrangian function of the optimization problem that is defined in (11) can be written in the following equation:

$$L(\theta, \mathbf{Z}, \mathbf{M}, \mathbf{h}, \mathbf{T}) = \frac{1}{b \times b \times d} \sum_{x=0}^{b-1} \sum_{y=0}^{b-1} \sum_{z=0}^{d-1} \left(X_{x,y,z} - X_{x,y,z}^{(M)}\right)^2 + \frac{\alpha}{2} \|\mathbf{W}\|_F^2$$
$$+ \lambda_1 \left\|\mathbf{X}^{(M_1)} - \mathbf{X}^{(M_1)} \mathbf{M}\right\|_F^2 + \lambda_1 \beta \|\mathbf{M}\|_1$$
$$+ \lambda_2 \sum_{x_i^{(M_1)} \in \mathcal{V}} -I(t) \log\left(p(\mathbf{y}_i / x_i^{(M_1)})\right) + \lambda_2 \eta \sum_{x_i^{(M_1)} \in \mathcal{V}} \sum_{x_j^{(M_1)} \in \mathcal{N}_k(x_i^{(M_1)})} -\mathbf{h}_2 \mathbf{S}_{i,j}$$
$$+ <\mathbf{T}, \mathbf{Z} - \mathbf{M}> + \frac{\varepsilon}{2} \|\mathbf{Z} - \mathbf{M}\|_F^2$$
$$= \frac{1}{b \times b \times d} \sum_{x=0}^{b-1} \sum_{y=0}^{b-1} \sum_{z=0}^{d-1} \left(X_{x,y,z} - X_{x,y,z}^{(M)}\right)^2 + \frac{\alpha}{2} \|\mathbf{W}\|_F^2$$
$$+ \lambda_1 \left\|\mathbf{X}^{(M_1)} - \mathbf{X}^{(M_1)} \mathbf{M}\right\|_F^2 + \lambda_1 \beta \|\mathbf{M}\|_1$$
$$+ \lambda_2 \sum_{x_i^{(M_1)} \in \mathcal{V}} -I(t) \log\left(p(\mathbf{y}_i / x_i^{(M_1)})\right) + \lambda_2 \eta \sum_{x_i^{(M_1)} \in \mathcal{V}} \sum_{x_j^{(M_1)} \in \mathcal{N}_k(x_i^{(M_1)})} -\mathbf{h}_2 \mathbf{S}_{i,j}$$
$$+ \frac{\varepsilon}{2} \left(\left\|\mathbf{Z} - \mathbf{M} + \frac{\mathbf{T}}{\varepsilon}\right\|_F^2\right) - \frac{1}{2\varepsilon} \left(\|\mathbf{T}\|_F^2\right)$$
(12)

where $\mathbf{T}$ is the Lagrangian multiplier, and $\varepsilon > 0$ is a penalty parameter.

$\theta$ can be optimized when $\mathbf{Z}$, $\mathbf{M}$ and $\mathbf{h}$ are fixed. The problem that is defined in (11) can be further rewritten as (13). For $\theta$, we can optimize the following problem:

$$\min_\theta L_1(\theta) = \min_\theta \left\{ \begin{array}{l} \frac{1}{b \times b \times d} \sum_{x=0}^{b-1} \sum_{y=0}^{b-1} \sum_{z=0}^{d-1} \left(X_{x,y,z} - X_{x,y,z}^{(M)}\right)^2 + \frac{\alpha}{2} \|\mathbf{W}\|_F^2 \\ + \lambda_1 \left\|\mathbf{X}^{(M_1)} - \mathbf{X}^{(M_1)}\mathbf{Z}\right\|_F^2 + \lambda_1 \beta \|\mathbf{Z}\|_1 \\ + \frac{\varepsilon}{2} \left( \left\|\mathbf{Z} - \mathbf{M} + \frac{\mathbf{T}}{\varepsilon}\right\|_F^2 \right) - \frac{1}{2\varepsilon}\left(\|\mathbf{T}\|_F^2\right) \end{array} \right\}$$

(13)

$\mathbf{Z}$ is computed when $\theta$, $\mathbf{M}$ and $\mathbf{h}$ are fixed. As for $\mathbf{Z}$, we can optimize the following problem:

$$\min_\mathbf{Z} L_2(\mathbf{Z}) = \min_\mathbf{Z} \left\{ \begin{array}{l} \lambda_1 \left\|\mathbf{X}^{(M_1)} - \mathbf{X}^{(M_1)}\mathbf{Z}\right\|_F^2 + \lambda_2 \eta \sum_{\mathbf{x}_i \in \mathcal{V}} \sum_{\mathbf{x}_j \in \mathcal{N}_k(\mathbf{x}_i)} \mathbf{h}_2 \mathbf{S}_{i,j} \\ + \frac{\varepsilon}{2} \|\mathbf{Z} - \mathbf{M} + \frac{\mathbf{T}}{\varepsilon}\|_F^2 \end{array} \right\}$$

(14)

$\mathbf{M}$ can be solved when $\theta$, $\mathbf{Z}$ and $\mathbf{h}$ are fixed. For the $\mathbf{M}$ subproblem, we can optimize the following problem:

$$\min_\mathbf{M} L_3(\mathbf{M}) = \min_\mathbf{M} \beta \|\mathbf{M}\|_1 + \frac{\varepsilon}{2} \|\mathbf{Z} - \mathbf{M} + \frac{\mathbf{T}}{\varepsilon}\|_F^2 \quad (15)$$

The Lagrangian multiplier $\mathbf{T}$ can be updated by

$$\mathbf{T}_{new} = \mathbf{T}_{old} + \varepsilon(\mathbf{Z} - \mathbf{M}) \quad (16)$$

For the $\mathbf{h}$ optimization, with the original objective function that is defined in (13), we can get the following equations:

$$\min_\mathbf{h} L_4(\mathbf{h}) = \min_\mathbf{h} \left\{ \begin{array}{l} \lambda_2 \sum_{\mathbf{x}_i^{(M_1)} \in \mathcal{V}} -I(t) \log\left(p(\mathbf{y}_i / \mathbf{x}_i^{(M_1)})\right) \\ + \lambda_2 \eta \sum_{\mathbf{x}_i^{(M_1)} \in \mathcal{V}} \sum_{\mathbf{x}_j^{(M_1)} \in \mathcal{N}_k\left(\mathbf{x}_i^{(M_1)}\right)} -\mathbf{h}_2 \mathbf{S}_{i,j} \end{array} \right\} \quad (17)$$

More details of SLCRF optimization can be referred to the Appendixes. In summary, we can form Algorithm 1 as follows.

---

**Algorithm 1:** SLCRF

**Input:** Training data $\mathbf{X}$, parameters $\beta, \gamma, \eta, \lambda_1, \lambda_2$.

**Initialize:** $\theta$, $\mathbf{Z}$, $\mathbf{h}$, $\mathbf{M}$, and $\mathbf{T}$; $\omega = 10^3, \varepsilon = 0.01, \alpha = 0.0005$, $\delta_1 = 0.001, \delta_2 = 1, \tau = 0.0002$.

**While** stopping criterion cannot meet **do**
1. Update $\theta$ using Eq. (13);
2. Update $\mathbf{Z}$ using Eq. (14);
3. Update auxiliary matrix $\mathbf{M}$ using Eq. (15);
4. Update $\mathbf{h}$ using Eq. (17);
5. Update the Lagrangian multiplier $\mathbf{T}$ using Eq. (16);

**Output:** The optimal solutions $\theta$, $\mathbf{Z}$, $\mathbf{h}$, $\mathbf{M}$, and $\mathbf{T}$.

---

## IV. EXPERIMENTS

In the experiment section, the performance of the proposed SLCRF approach for HSI classification is evaluated using different datasets. First, a brief description of the three used HSI datasets is presented, and then the classification results of the SLCRF with some related methods are further compared.

### A. Experimental Datasets

**Indian Pines dataset.** The AVIRIS sensor collected corresponding data at the Indian Pines site in 1992. The image size is approximately 145×145. The dataset consists of 224 spectral bands, and the spatial resolution is 20 m. The wavelength range is 0.4–2.5 μm. The bands of the water absorption region are discarded and 200 out of the 224 bands are kept in the experiment. The details of the Indian Pines dataset that covers 16 classes and 10,249 pixels are provided in Table II.

**University of Pavia dataset** (PaviaU). These data from northern Italy was acquired for an urban area using the ROSIS-03 sensor. The HSI size is approximately 610×340, while its spatial resolution is approximately 1.3 meters. In the experiment, 103 out of the 115 bands are used, and the water absorption bands are discarded. As a result, the details of the dataset that covers 9 classes and 42,776 labeled samples are provided in Table III.

**The Houston dataset** (2018). The Houston data was organized by the Image Analysis and Data Fusion Technical Committee. The size of HSI is approximately 601×2384 pixels and 48 spectral bands, while the spatial resolution is approximately 1.0 meters. The details of the Houston dataset that includes 20 categories and 504,712 labeled pixels are provided in Table IV.

5% training samples for PaviaU and Houston dataset and 10% training samples for Indian pines dataset are shown in Tables II-IV. To evaluate our proposed method compared to the related methods, the HSIs are classified under different percentages of training pixels, including 5, 10, 15 and 20.

TABLE II
THE INDIAN PINES DATASET

| Class | Land Cover Type | Training (10%) | Labeling (5%) | Total |
|---|---|---|---|---|
| 1 | Alfalfa | 5 | 2 | 46 |
| 2 | Corn-notill | 143 | 71 | 1,428 |
| 3 | Corn-mintill | 83 | 42 | 830 |
| 4 | Corn | 24 | 12 | 237 |
| 5 | Grass-pasture | 48 | 24 | 483 |
| 6 | Grass-trees | 73 | 36 | 730 |
| 7 | Grass-pasture-mowed | 3 | 1 | 28 |
| 8 | Hay-windrowed | 48 | 24 | 478 |
| 9 | Oats | 2 | 1 | 20 |
| 10 | Soybean-notill | 97 | 49 | 972 |
| 11 | Soybean-mintill | 245 | 123 | 2,455 |
| 12 | Soybean-clean | 59 | 29 | 593 |
| 13 | Wheat | 20 | 10 | 205 |
| 14 | Woods | 126 | 63 | 1,265 |
| 15 | Bldg-grass-trees | 39 | 20 | 386 |
| 16 | Stone-Steel-Towers | 9 | 5 | 93 |
| | Total | 1,024 | 512 | 10,249 |

TABLE III
THE PAVIAU DATASET

| Class | Land Cover Type | Training (5%) | Labeling (1%) | Total |
|---|---|---|---|---|
| 1 | Asphalt | 345 | 66 | 6,631 |
| 2 | Meadows | 917 | 186 | 18,649 |
| 3 | Gravel | 131 | 20 | 2,099 |
| 4 | Trees | 156 | 30 | 3,064 |
| 5 | Metal sheets | 77 | 13 | 1,345 |
| 6 | Bare Soil | 246 | 50 | 5,029 |
| 7 | Bitumen | 60 | 13 | 1,330 |
| 8 | Bricks | 181 | 36 | 3,682 |
| 9 | Shadows | 44 | 94 | 947 |

| | | Total | 2,157 | 508 | 42,776 |
|---|---|---|---|---|---|

TABLE IV
THE HOUSTON DATASET

| Class | Land Cover Type | Training (5%) | Labeling (1%) | Total |
|---|---|---|---|---|
| 1 | Healthy Grass | 490 | 98 | 9,799 |
| 2 | Stressed Grass | 1,625 | 325 | 32,502 |
| 3 | Artificial Turf | 34 | 7 | 684 |
| 4 | Evergreen Trees | 679 | 136 | 13,588 |
| 5 | Deciduous Trees | 252 | 50 | 5,048 |
| 6 | Bare Earth | 225 | 45 | 4,516 |
| 7 | Water | 13 | 2 | 266 |
| 8 | Residential Buildings | 1,988 | 397 | 39,762 |
| 9 | Non-residential Buildings | 11,184 | 2,237 | 223,684 |
| 10 | Roads | 2,290 | 458 | 45,810 |
| 11 | Sidewalks | 1,700 | 340 | 34,002 |
| 12 | Crosswalks | 75 | 15 | 1,516 |
| 13 | Major Thoroughfares | 2,318 | 463 | 46,358 |
| 14 | Highways | 492 | 98 | 9,849 |
| 15 | Railways | 347 | 69 | 6,937 |
| 16 | Paved Parking Lots | 574 | 115 | 11,475 |
| 17 | Unpaved Parking Lots | 7 | 1 | 149 |
| 18 | Cars | 333 | 65 | 6,578 |
| 19 | Trains | 267 | 53 | 5,365 |
| 20 | Stadium Seats | 346 | 68 | 6,824 |
| | Total | 25,239 | 5,042 | 504,712 |

*B. Related Approaches*

The developed method SLCRF is semi-supervised for the relationship construction between the pixels in HSIs. The following approaches are adopted to compare with the SLCRF with respect to the accuracy of the HSI classification.

1) PCA and softmax classifier (PCA-SC). First, the PCA is adopted to reduce the dimensionality of the spectral dimension of HSI, and the neighborhood region of the pixel in the condensed data is extracted and flattened into a vector. The softmax classifier is used to classify the flattened vector with the labeled data.

2) Locality preserving projections [32] and SC (LPP-SC). LPP is used to map the HSI into a low dimension. The neighborhood region of the pixel in the low dimension is fed into a two-layer CNN. The softmax classifier is used to classify the spatial-spectral features with the labeled data.

3) Convolutional autoencoder [33] and SC (CAE-SC). CAE is a network that can be used to learn features of HSI in an unsupervised way. The CAE consists of one encoder function and one decoder function, and the encoder and decoder each have one convolutional layer. The encoder function maps the spatial-spectral input to a new feature representation of input data. Then, SC is used to classify the features extracted by CAE with the labeled data.

4) Stacted CAE [30] and SC (SCAE-SC). The SCAE is a network that consists of two shallow CAE. The outputs of the hidden layer of CAE are wired to the input of subsequent CAE. The softmax classifier classifies the spatial-spectral features with the labeled data.

5) 3D Convolutional Autoencoder [29] and CRF (3DCAE-CRF). The operations of 3DCAE are all 3D convolution, 3D pooling, and 3D batch normalization to explore spatial-spectral information. The features learned by 3DCAE are flattened into a one-dimensional vector. The CRF is used to produce labels of the one-dimensional vector with the labeled data, which is implemented by using the GraphCRF and NSlackSSVM tools of the PyStruct.

Moreover, to obtain the best HSI classifications, the following settings need to be further assessed. For the three datasets, *s* training samples are randomly selected as the labeled data. For the Indian pines dataset, the labeled data is set 5%, 6%, 7%, 8%, 9%, and 10% for each class, and for the PaviaU and Houston dataset, the labeled data is set 1%, 2%, 3%, 4% and 5% for each class. The remaining HSI data are adopted for the classification.

In our SLCRF method, it selects *k* from the set {3, 4, 5, 6, 7, 8, 9, 10} to construct the spatial graph. The balancing parameters such as $\lambda_1$ and $\lambda_2$ are tuned using the set {$10^{-9}$, $10^{-8}$, …, $10^8$, $10^9$}.

In the CAE, SCAE, 3DCAE, and our SLCRF methods, the number of feature dimensions is selected from the set {*d*/4, *d*/3, *d*/2, *d*, *d*×2, *d*×3, *d*×4}, where *d* is the number of spectrum bands. The number of the batch size is selected from the set {16, 32, 64, 128, 256}. The learning rate is selected from the set {0.01, 0.001, 0.0001, 0.0002, 0.00001, 0.00005}. The main architectures of SLCRF for the three datasets are shown in Tables V-VII.

TABLE V
ARCHITECTURES OF SLCRF FOR INDIAN PINES DATASET

| | Layer | Convolution | Srides | BN | Activation |
|---|---|---|---|---|---|
| Encoder | Conv1 | 24×3×3×24 | 1×1×1 | Yes | ReLU |
| | Conv2 | 24×3×3×48 | 20×1×1 | Yes | ReLU |
| | FC1 | 432×216 | 1 | No | ReLU |
| | FC2 | 216×144 | 1 | No | ReLU |
| Decoder | FC3 | 144×216 | 1 | No | ReLU |
| | FC4 | 216×432 | 1 | No | ReLU |
| | Deconv1 | 9×3×3×24 | 22 | Yes | ReLU |
| | Deconv2 | 27×3×3×24 | 1 | Yes | ReLU |

*BN denotes the batch normalization, and FC denotes fully connection.*

TABLE VI
ARCHITECTURES OF SLCRF FOR PAVIAU DATASETS

| | Layer | Convolution | Srides | BN | Activation |
|---|---|---|---|---|---|
| Encoder | Conv1 | 11×3×3×24 | 1 | Yes | ReLU |
| | Conv2 | 11×3×3×48 | 1 | Yes | ReLU |
| | Max.Pool | 9×1×1 | 9 | No | NO |
| | FC1 | 432×216 | 1 | No | ReLU |
| Decoder | FC2 | 216×432 | 1 | No | ReLU |
| | Deconv1 | 9×3×3×24 | 10 | Yes | ReLU |
| | Deconv2 | 9×3×3×24 | 1 | Yes | ReLU |

*BN denotes the batch normalization, and FC denotes fully connection.*

TABLE VII
ARCHITECTURES OF SLCRF FOR HOUSTON DATASETS

| | Layer | Convolution | Srides | BN | Activation |
|---|---|---|---|---|---|
| Encoder | Conv1 | 5×3×3×24 | 1 | Yes | ReLU |
| | Conv2 | 5×3×3×48 | 1 | Yes | ReLU |
| | Max.Pool | 4×1×1 | 9 | No | NO |
| | FC1 | 480×240 | 1 | No | ReLU |
| Decoder | FC2 | 240×480 | 1 | No | ReLU |
| | Deconv1 | 4×3×3×24 | 4 | Yes | ReLU |
| | Deconv2 | 9×3×3×24 | 1 | Yes | ReLU |

*BN denotes the batch normalization, and FC denotes fully connection.*

## C. Evaluation Results

The HSI classification performances of the different methods are computed and compared in Tables VIII -XI and Figs. 2-5 using three metrics such as the overall accuracy (OA), the average accuracy (AA), as well as the Kappa coefficient.

We can further obtain the following observations from Tables VIII -XI and Figs. 2-5.

1) Compared with the related methods including the PCA-SC, LPP-SC, CAE-SC, SCAE-SC and 3DCAE-CRF, the SLCRF can obtain the best HSI classification results for the testing image samples using the three datasets, which are much better than those that are acquired by other methods. The results further show that the SLCRF very appropriate for image classification.
2) From Figs. 2-4, the HSI classification results that are achieved by the SLCRF using the three datasets are more compact than the related methods. The main reason is that the relationships between image pixels can be well built by the SLCRF. These good relationships are very beneficial for image classification.
3) In Fig. 5, as the number of the labeled image pixels increases, the image classification accuracies of all approaches are improved. However, the SLCRF can obtain more accurate HSI overall classifications using different numbers of labeled image samples, which further demonstrates that the SLCRF can outperform the other methods. In the proposed SLCRF method, the SL, the relationship construction and the semi-supervised constraints that are defined in the CRF are integrated together to construct a unified objective function. The relationships between the image pixels in HSIs can be better described using the constructed spectral-spatial graph. Moreover, the CRF framework can enhance the performance of SL. Therefore, the SLCRF obtains much better HSI classification results.
4) In Table. XI, the effects of training samples are analyzed by comparing the OA performance of SLCRF with CAE-SC as the percentage of the training dataset is changed: 5%, 10%, 15%, and 20%. The classification accuracy of the proposed SLCRF method increases as the training samples increases. SLCRF provides higher accuracy than CAE-SC using different numbers of training samples.

TABLE XI
PERFORMANCE COMPARISON WITH DIFFERENT PERCENTAGES TRAINING SAMPLES WITH LABELED SAMPLES

| Dataset | Method | 5% | 10% | 15% | 20% |
|---|---|---|---|---|---|
| Indian pines | CAE-SC | 72.62 | 72.66 | 73.49 | 75.93 |
| | SLCRF | **83.24** | **90.70** | **91.22** | **92.84** |
| PaviaU | CAE-SC | 90.27 | 91.17 | 89.88 | 90.68 |
| | SLCRF | **94.48** | **96.26** | **97.18** | **97.59** |
| Houston | CAE-SC | 80.05 | 80.78 | 81.65 | 83.55 |
| | SLCRF | **85.64** | **87.61** | **87.91** | **88.43** |

TABLE VIII
SEMI-SUPERVISED CLASSIFICATION RESULTS (%) WITH 5% LABELED TRAINING SAMPLES FOR THE INDIAN PINES DATASET

| Class | PCA-SC | LPP-SC | CAE-SC | SCAE-SC | 3DCAE-CRF | SLCRF |
|---|---|---|---|---|---|---|
| Alfalfa | 9.30 | 20.93 | 8.89 | 0.00 | 49.05 | 85.71 |
| Corn-notill | 66.52 | 49.04 | 66.54 | 73.59 | 75.18 | 90.03 |
| Corn-mintill | 50.38 | 51.90 | 57.63 | 53.18 | 70.51 | 87.33 |
| Corn | 30.67 | 10.22 | 26.32 | 32.44 | 65.00 | 81.65 |
| Grass-pasture | 78.17 | 87.55 | 75.65 | 88.08 | 64.67 | 92.22 |
| Grass-trees | 95.82 | 86.15 | 96.22 | 95.09 | 89.53 | 95.16 |
| Grass-pasture-mowed | 11.54 | 19.23 | 0.00 | 11.11 | 27.54 | 54.55 |
| Hay-windrowed | 100.00 | 96.92 | 98.66 | 99.56 | 95.71 | 98.85 |
| Oats | 21.05 | 15.79 | 31.58 | 72.22 | 29.27 | 72.22 |
| Soybean-notill | 57.20 | 53.52 | 64.60 | 68.39 | 79.91 | 86.93 |
| Soybean-mintill | 79.50 | 80.92 | 77.46 | 79.93 | 85.45 | 91.42 |
| Soybean-clean | 22.20 | 29.66 | 41.52 | 38.74 | 75.38 | 83.62 |
| Wheat | 100.00 | 87.11 | 97.47 | 86.32 | 72.02 | 96.81 |
| Woods | 95.17 | 91.01 | 94.95 | 94.52 | 92.22 | 96.41 |
| Bldg-grass-trees | 54.64 | 55.74 | 40.55 | 48.79 | 63.12 | 81.63 |
| Stone-Steel-Towers | 78.41 | 86.36 | 82.95 | 82.23 | 90.00 | 95.40 |
| OA | 71.62 | 68.27 | 72.66 | 74.80 | 79.62 | **90.70** |
| AA | 59.41 | 55.63 | 60.06 | 63.89 | 70.28 | **86.87** |
| Kappa | 0.671 | 0.634 | 0.686 | 0.710 | 0.768 | **0.894** |

*The best results are highlighted in bold.*

TABLE IX
SEMI-SUPERVISED CLASSIFICATION RESULTS (%) WITH 1% LABELED TRAINING SAMPLES FOR THE PAVIAU DATASET

| Class | PCA-SC | LPP-SC | CAE-SC | SCAE-SC | 3DCAE-CRF | SLCRF |
|---|---|---|---|---|---|---|
| Asphalt | 90.45 | 81.67 | 94.09 | 89.49 | 81.79 | 95.29 |
| Meadows | 96.42 | 93.81 | 96.98 | 96.95 | 91.78 | 97.83 |
| Gravel | 79.59 | 62.25 | 78.25 | 78.30 | 83.20 | 86.52 |
| Trees | 83.25 | 78.04 | 88.43 | 81.59 | 86.44 | 94.26 |
| Metal sheets | 99.47 | 92.19 | 93.54 | 99.62 | 73.99 | 99.78 |
| Bare Soil | 73.50 | 85.66 | 79.93 | 74.80 | 73.90 | 85.09 |
| Bitumen | 76.37 | 91.11 | 80.09 | 80.85 | 79.07 | 82.93 |
| Bricks | 82.69 | 80.63 | 72.34 | 87.25 | 78.24 | 92.03 |
| Shadows | 98.93 | 54.64 | 98.08 | 96.59 | 85.58 | 97.36 |
| OA | 89.38 | 86.30 | 90.27 | 89.91 | 85.06 | **94.48** |
| AA | 86.74 | 80.33 | 86.86 | 87.27 | 81.55 | **92.68** |
| Kappa | 0.858 | 0.817 | 0.870 | 0.865 | 0.803 | **0.926** |

*The best results are highlighted in bold.*

TABLE X
SEMI-SUPERVISED CLASSIFICATION RESULTS (%) WITH 1% LABELED TRAINING SAMPLES FOR THE HOUSTON DATASET

| Class | PCA-SC | LPP-SC | CAE-SC | SCAE-SC | 3DCAE-CRF | SLCRF |
|---|---|---|---|---|---|---|
| Healthy Grass | 83.70 | 70.62 | 91.21 | 94.61 | 83.99 | 89.14 |
| Stressed Grass | 92.66 | 77.14 | 93.80 | 93.22 | 89.08 | 94.46 |
| Artificial Turf | 99.56 | 96.60 | 98.52 | 99.41 | 67.21 | 96.61 |
| Evergreen Trees | 94.96 | 64.12 | 94.93 | 93.62 | 82.75 | 95.35 |
| Deciduous Trees | 56.01 | 41.22 | 60.58 | 64.88 | 65.40 | 73.22 |
| Bare Earth | 54.11 | 70.60 | 83.58 | 85.86 | 55.21 | 92.03 |
| Water | 90.11 | 66.54 | 89.35 | 91.63 | 30.46 | 87.10 |
| Residential Buildings | 71.04 | 71.72 | 76.98 | 74.44 | 61.08 | 82.12 |
| Non-residential Buildings | 90.45 | 89.45 | 93.30 | 92.92 | 86.23 | 94.78 |
| Roads | 45.34 | 49.32 | 48.24 | 44.63 | 58.58 | 63.66 |
| Sidewalks | 44.00 | 22.10 | 51.22 | 47.13 | 53.62 | 59.46 |
| Crosswalks | 0.00 | 0.13 | 0.33 | 0.00 | 0.00 | 15.02 |
| Major Thoroughfares | 50.51 | 61.97 | 65.43 | 73.72 | 56.49 | 77.69 |
| Highways | 45.36 | 74.29 | 69.34 | 78.61 | 51.28 | 88.29 |
| Railways | 79.67 | 85.51 | 93.84 | 96.51 | 74.79 | 98.10 |
| Paved Parking Lots | 72.02 | 75.18 | 77.44 | 84.13 | 59.69 | 86.12 |
| Unpaved Parking Lots | 0.00 | 29.93 | 31.29 | 2.72 | 2.83 | 75.56 |
| Cars | 36.96 | 46.45 | 38.30 | 43.96 | 62.49 | 64.53 |
| Trains | 69.63 | 61.74 | 74.64 | 77.44 | 61.20 | 87.53 |
| Stadium Seats | 85.05 | 88.82 | 87.70 | 89.42 | 70.96 | 93.17 |
| OA | 74.77 | 73.03 | 80.05 | 80.38 | 74.56 | **85.64** |
| AA | 63.06 | 62.17 | 71.00 | 71.44 | 58.67 | **80.70** |
| Kappa | 0.667 | 0.645 | 0.739 | 0.744 | 0.667 | **0.813** |

*The best results are highlighted in bold.*

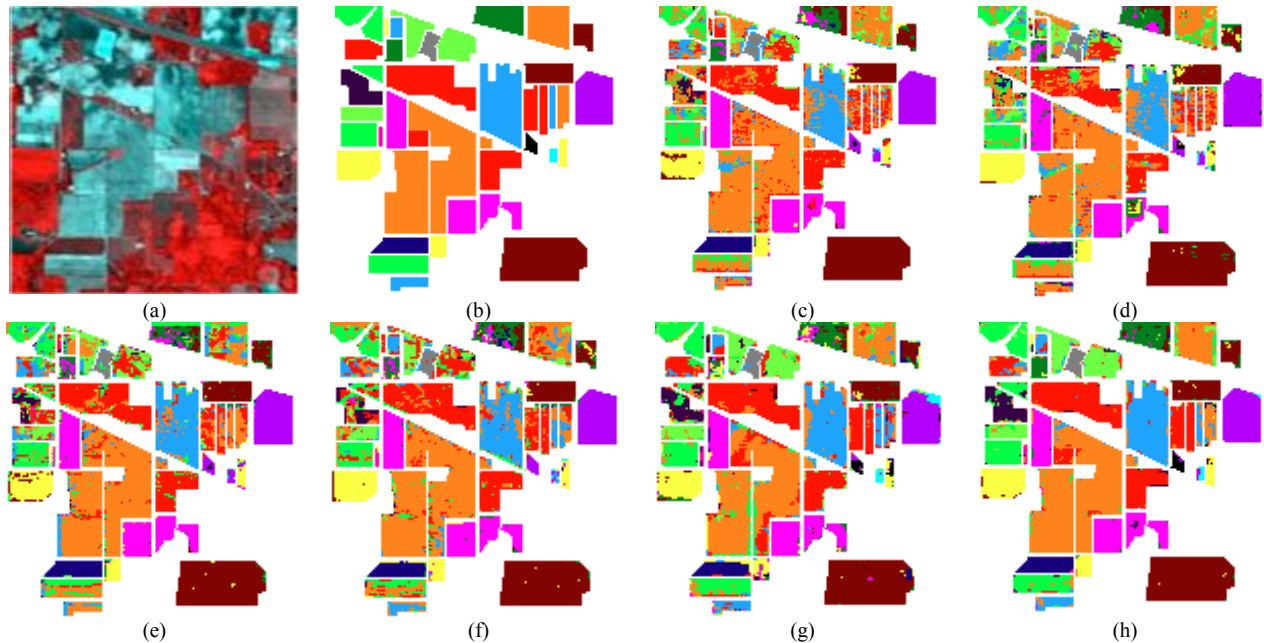

(a) (b) (c) (d)
(e) (f) (g) (h)

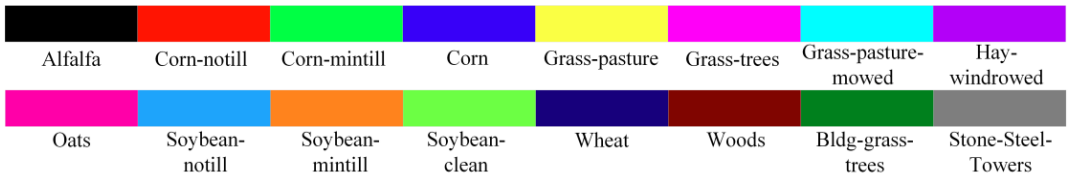

Fig. 2. Classification maps of the different methods with 5% labeled samples for the Indian pines dataset. (a) False-color. (b) Ground truth. (c) PCA-SC. (d) LPP-SC. (e) CAE-SC. (f) SCAE-SC. (g) 3DCAE-CRF. (h) SLCRF.

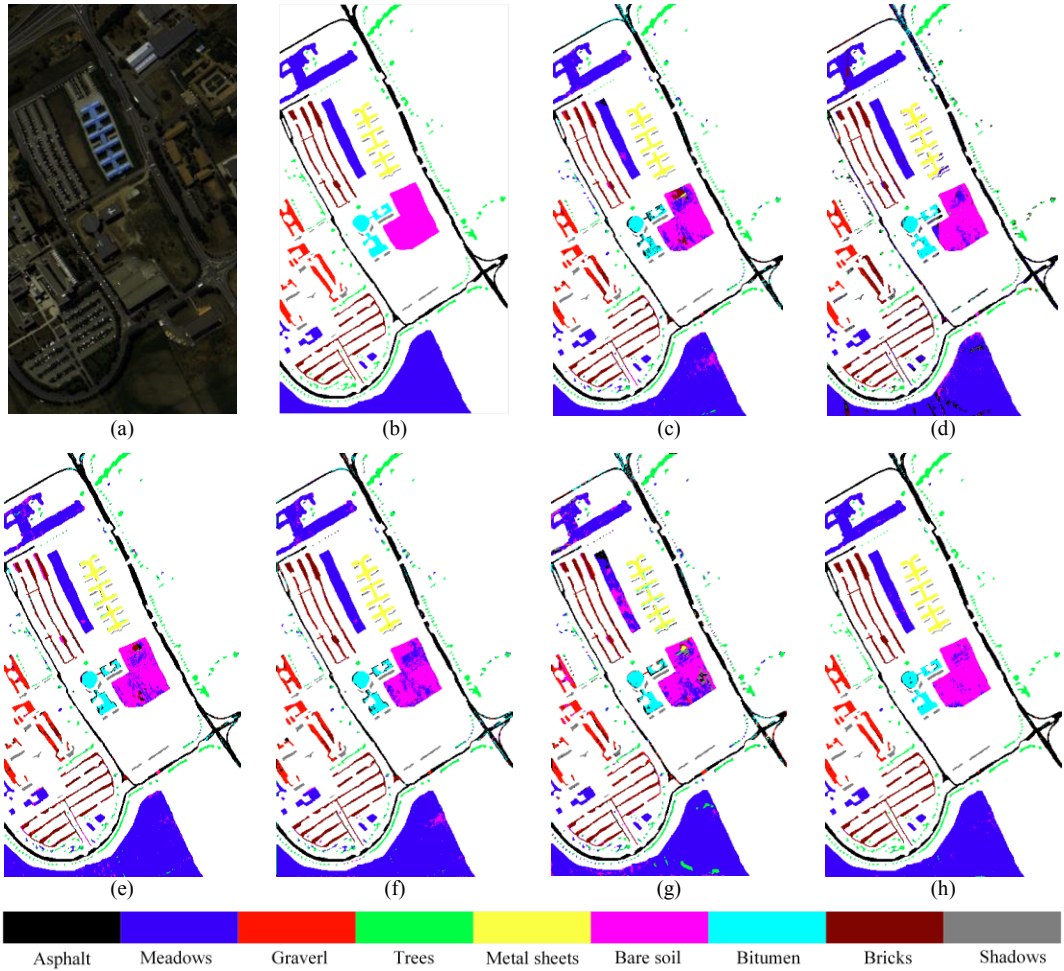

Fig. 3. Classification maps of the different methods with 1% percentage labeled samples for the PaviaU dataset. (a) False-color. (b) Ground truth. (c) PCA-SC. (d) LPP-SC. (e) CAE-SC. (f) SCAE-SC. (g) 3DCAE-CRF. (h) SLCRF.

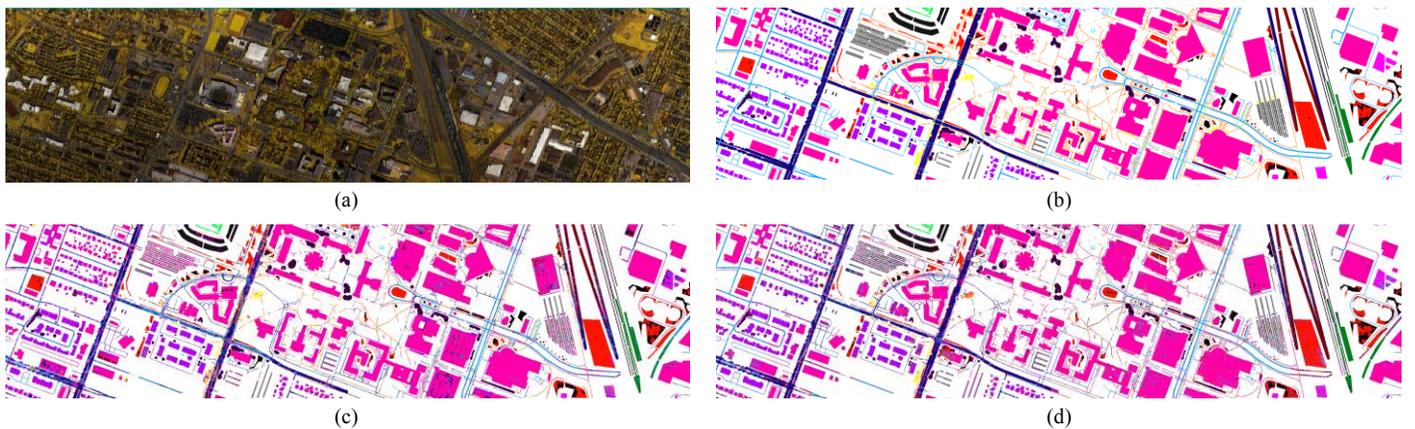

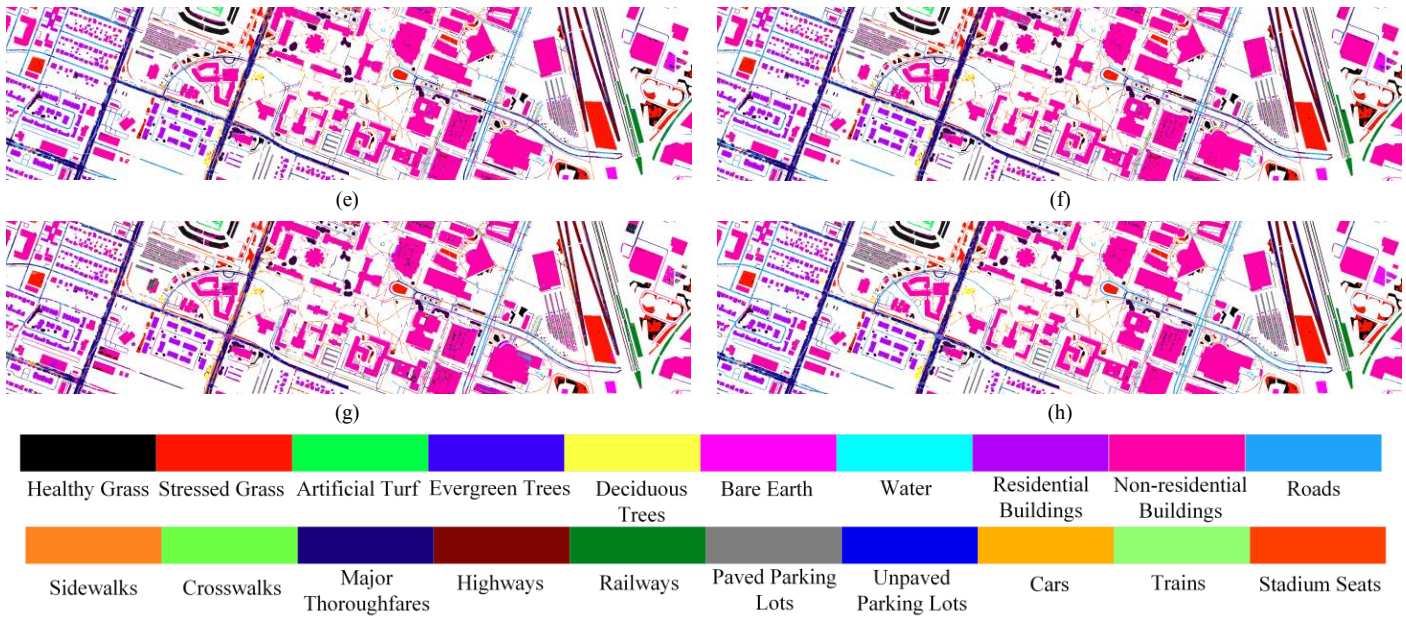

(e) (f)

(g) (h)

Healthy Grass | Stressed Grass | Artificial Turf | Evergreen Trees | Deciduous Trees | Bare Earth | Water | Residential Buildings | Non-residential Buildings | Roads

Sidewalks | Crosswalks | Major Thoroughfares | Highways | Railways | Paved Parking Lots | Unpaved Parking Lots | Cars | Trains | Stadium Seats

Fig. 4. Classification maps of the different methods with 1% labeled samples for the Houston dataset. (a) False-color. (b) Ground truth. (c) PCA-SC. (d) LPP-SC. (e) CAE-SC. (f) SCAE-SC. (g) 3DCAE-CRF. (h) SLCRF.

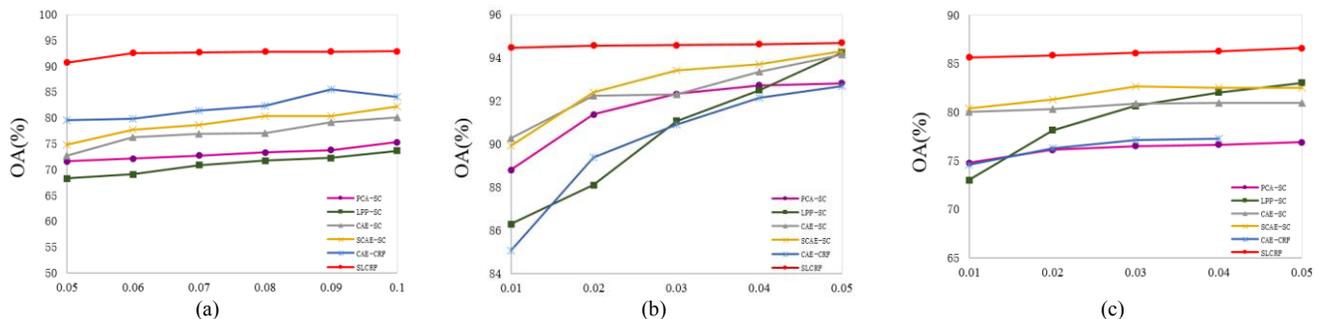

(a) (b) (c)

Fig. 5. Classification results of the PCA-SC, LPP-SC, CAE-SC, SCAE-SC, 3DCAE-CRF and SLCRF with different percentages of labeled image samples. (a) The Indian pines dataset. (b) The PaviaU dataset. (c) The Houston dataset.

### D. Effectiveness of the Relationships' Construction Term

To evaluate the effectiveness of the relationships' construction term in the proposed method, we compare the SLCRF with the first reconstruction term and the CRF term, called RE-CRF, and is defined with the following equations:

$$\min_{\theta, \mathbf{Z}, \mathbf{h}} \frac{1}{b \times b \times d} \sum_{x=0}^{b-1} \sum_{y=0}^{b-1} \sum_{z=0}^{d-1} \left(X_{x,y,z} - X_{x,y,z}^{(M)}\right)^2 + \frac{\alpha}{2} \|\mathbf{W}\|_F^2$$
$$+ \lambda_2 \sum_{\mathbf{x}_i^{(M_1)} \in \mathcal{V}} -I(t) \log\left(p(\mathbf{y}_i / \mathbf{x}_i^{(M_1)})\right) \quad (18)$$
$$+ \lambda_2 \eta \sum_{\mathbf{x}_i^{(M_1)} \in \mathcal{V}} \sum_{\mathbf{x}_j^{(M_1)} \in \mathcal{N}_k\left(\mathbf{x}_i^{(M_1)}\right)} -\mathbf{h}_2 \mathbf{S}_{2i,j}$$

From Table XII, the performance of the SLCRF significantly outperforms the method without the relationships' construction term in the three datasets, which further demonstrate that the relationships' construction term can provide a better solution to enhance the relationships of pixels.

TABLE XII
CLASSIFICATION RESULTS (%) WITH LABELED IMAGE SAMPLES IN EACH CLASS ON THE THREE DATASETS.

| Dataset | RE-CRF | SLCRF |
|---|---|---|
| Indian pines | 86.02 | **90.70** |
| PaviaU | 89.92 | **94.48** |
| Houston | 82.72 | **85.64** |

*The best results are highlighted in bold.*

### E. Effectiveness of the CRF Term

To evaluate the effectiveness of the CRF in the proposed method, we compare the SLCRF with SL whose model does not consider the CRF framework and is defined with the following equations:

$$\min_{\theta, \mathbf{Z}} \frac{1}{b \times b \times d} \sum_{x=0}^{b-1} \sum_{y=0}^{b-1} \sum_{z=0}^{d-1} \left(X_{x,y,z} - X_{x,y,z}^{(M)}\right)^2 + \frac{\alpha}{2} \|\mathbf{W}\|_F^2 \quad (19)$$
$$+ \lambda_1 \left\|\mathbf{X}^{(M_1)} - \mathbf{X}^{(M_1)} \mathbf{Z}\right\|_F^2 + \lambda_1 \beta \|\mathbf{Z}\|_1$$

The features extracted by the SL are classified by using random forest classifier (SL-RF). From the HSI classification results in Table XIII, we can see that the SLCRF significantly outperforms the model without the CRF in the three datasets. These results can further validate the fact that CRF can provide a better solution to enhance the performance of SL and incorporate the features for visual recognition.

TABLE XIII
CLASSIFICATION RESULTS (%) WITH LABELED IMAGE SAMPLES IN EACH CLASS ON THE THREE DATASETS.

| Dataset | SL-RF | SLCRF |
|---|---|---|
| Indian pines | 87.46 | **90.70** |
| PaviaU | 90.67 | **94.48** |
| Houston | 83.24 | **85.64** |

*The best results are highlighted in bold.*

### F. Parameters Analysis

In the proposed method SLCRF, five parameters including $\beta$, $\gamma$, $\eta$, $\lambda_1$ and $\lambda_2$ need to be tuned in each dataset. The influences of these parameters on the HSI classification results are discussed.

The parameters $\lambda_1$ and $\lambda_2$ correspond to the pixels' relationships construction and CRF regularization, respectively. We have tuned these parameters as shown in Fig. 6. Since every dataset has its own distinctive data structure, the suitable parameters are different for each dataset. From Fig. 6, it is also seen that the values of parameters $\gamma$ and $\eta$ have greater impacts on the image classification results than parameters $\beta$. The most suitable values of these parameters can be taken from the image classification performance. In conclusion, the best image classification results for the Indian pines dataset are achieved when the five parameters $\beta$, $\gamma$, $\eta$, $\lambda_1$ and $\lambda_2$ are close to $10^2$, $10$, $10^4$, $10^3$ and $10^{-3}$, respectively. For the PaviaU dataset, the best six parameters are those that are close to $10^3$, $10^2$, $10^2$, $10^5$, $10^3$ and $10^{-9}$, respectively. For the Houston dataset, the best $\beta$, $\gamma$, $\eta$, $\lambda_1$ and $\lambda_2$ are be $10^3$, $10$, $10$, $10^3$ and $10^{-9}$, respectively.

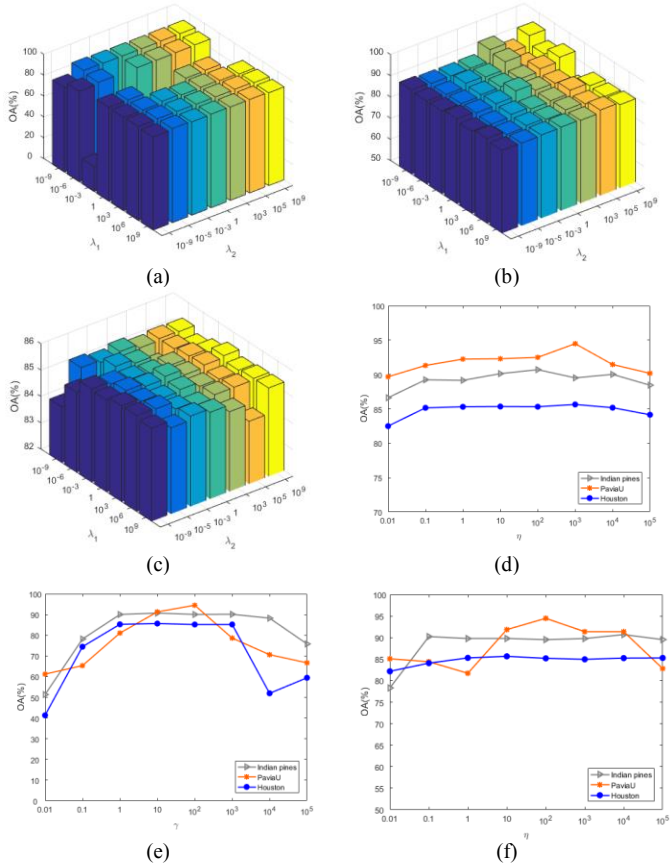

Fig. 6. Influences of the different parameter values on the HSI classification results. (a)-(c): The influences of $\lambda_1$ and $\lambda_2$ on the classification results for the Indian pines, PaviaU and Houston datasets, respectively. (d)-(f): The influences of different values of $\beta$, $\gamma$ and $\eta$ on the classification results for Indian pines, PaviaU and Houston.

## V. CONCLUSIONS

In this paper, a novel SL approach termed SLCRF is developed for the HSI classification. The main contributions of the SLCRF are its accurate integration of the probabilities of pixels belonging to different categories and the relationships between the neighboring pixels into the terms of the CRF. In this way, SLCRF can adopt semi-supervised learning to alleviate the need for labeled samples. The proposed SLCRF method integrates the SL, the relationship construction and the semi-supervised constraints that are defined in the CRF to build our objective function. The SLCRF uses an iterative optimization algorithm. Our proposed method is evaluated on three publicly used HSI classification datasets, and the experimental results show that the SLCRF method can outperform the related semi-supervised HSI classification approaches.

In future work, SL will be integrated in other deep learning frameworks, such as graph convolutional network, to automatically obtain more discriminative and representative features of the HSI and further boost the performance of the HSI classification.

## APPENDIX

### A. Operations for CAE

*1) 3D Convolutional Layer:* The 3D convolution kernel in the $m$-th convolution layer has the size of $P^{(m)} \times Q^{(m)} \times R^{(m)}$, where $R^{(m)}$ is the size of kernel along the spectral dimension. The value at position $(x, y, z)$ on the $j$-th feature map in the $m$-th layer ($1 \leq m < M_1 - l$) $X^{(m)}_{x,y,z}$ can be described as:

$$X^{(m,j)}_{x,y,z} = \varphi\left(b^{(m,j)} + \sum_{k=0}^{K^{(m)}-1}\sum_{p=0}^{P^{(m)}-1}\sum_{q=0}^{Q^{(m)}-1}\sum_{r=0}^{R^{(m)}-1} W^{(m,j,k)}_{p,q,r} X^{(m-1,k)}_{(x+p)(y+q)(z+r)}\right) \quad (20)$$

where $\varphi$ is the activation function. $j$ is the number of kernels in this layer, and $K^{(m)}$ is the number of feature cubes input to the $m$-th layer. $W^{m,j,n}_{p,q,r}$ is the $(p, q, r)$-th value of the kernel that is connected to the $k$-th feature map in the previous layer, and $b_{m,j}$ is the bias.

*2) 3D Max Pooling:* The size of the feature map extracted by the convolutional layer is reduced using the form of pooling. Max-pooling is the procedure that outputs the max value within the kernel size. For a $s \times s \times v$ window size neighbor denoted as $\Omega$, the output of max-pooling can be described as follows:

$$X^{(m)}_{x,y,z} = \max_{(x,y,z)\in\Omega} X^{(m-1)}_{x,y,z} \quad (21)$$

where $X^{(m-1)}_{x,y,z}$ ($1 \leq m < M_1 - l$) represents the features extracted from the 3D convolutional layer, and $X^{(m)}_{x,y,z}$ represents the features at position $(x, y, z)$ after 3D max pooling.

*3) 3D Deconvolution Layer:* The down-sampled feature maps are upsampled through the deconvolution layer to map into a high-dimensional space. Compared to simple upsampling, the convolution is a trainable upsampling convolutional layer, which can maintain the important details. Let $\mathbf{X}^{(m-1)'}$ be the input of $m$-th layer ($M_1 + l < m \leq M$) that have been padded with zeros in all the three dimensions. Then, the output of 3D deconvolution layer is obtained by employing the convolutional kernels on $\mathbf{X}^{(m-1)'}$.

$$X_{x,y,z}^{(m,j)} = \varphi \left[ b^{(m,j)} + \sum_{k=0}^{K^{(m)}-1} \sum_{p=0}^{P^{(m)}-1} \sum_{q=0}^{Q^{(m)}-1} \sum_{r=0}^{R^{(m)}-1} W_{p,q,r}^{(m,j,k)} \ X_{(x+p)(y+q)(z+r)}^{(m-1,k)} \right]' \quad (22)$$

*4) Fully Connected Layer:* At the last 3D convolution layer, the feature map of the previous layer are flattened and fed into fully connected layers. There are $2l$ fully connected layers in CAE, where the encoder and the decoder each has $l$ ($1 < l < M_1$) layers.

$$X_{x,y,z}^{(m)} = \sum_{i=1}^{L^{(m)}} \varphi \ b_i^{(m)} + W_i^{(m)} X_{x,y,z}^{(m-1)} \quad (23)$$

where $L^{(m)}$ is the number of the hidden nodes at the *m*-th layer ($M_1 - l \le m \le M_1 + l$). $W_i^{(m)}$ is the weights connecting the *i*-th hidden node and the input node, and $b_i^{(m)}$ is the bias of the *i*-th hidden node.

### B. Optimization for SLCRF

*1) Optimization for $\theta$:* $\theta$ is solved when $\mathbf{Z}$, $\mathbf{M}$, and $\mathbf{h}$ are fixed. We can optimize the following problem:

$$\min_{\theta} L_1(\theta) = \min_{\theta} \left\{ \begin{array}{l} \frac{1}{b \times b \times d} \sum_{x=0}^{b-1}\sum_{y=0}^{b-1}\sum_{z=0}^{d-1} \left( X_{x,y,z} - X_{x,y,z}^{(M)} \right)^2 \\ + \frac{\alpha}{2} \|\mathbf{W}\|_F^2 + \lambda_1 \|\mathbf{X}^{(M_1)} - \mathbf{X}^{(M_1)}\mathbf{Z}\|_F^2 + \lambda_1 \beta \|\mathbf{Z}\|_1 \\ + \frac{\varepsilon}{2}\left( \left\| \mathbf{Z} - \mathbf{M} + \frac{\mathbf{T}}{\varepsilon} \right\|_F^2 \right) - \frac{1}{2\varepsilon}\left( \|\mathbf{T}\|_F^2 \right) \end{array} \right\} \quad (24)$$

The optimization problem in (24) is solved using stochastic gradient descent and back-propagation algorithm, iteratively. The partial derivative for parameters of the *m*-th layer ($1 \le m \le M$) of (24) is

$$\frac{\partial L_1(\theta)}{\partial \mathbf{W}^{(m)}} = \mathbf{U}^{(m)} (\mathbf{X}^{(m-1)})^T + \alpha \sum_{k=0}^{K^{(m)}-1} \sum_{p=0}^{P^{(m)}-1} \sum_{q=0}^{Q^{(m)}-1} \sum_{r=0}^{R^{(m)}-1} W_{p,q,r}^{(m,j,k)} \quad (25)$$

The partial derivative for bias of the *m*-th layer of (24) is

$$\frac{\partial L_1(\theta)}{\partial \mathbf{b}^{(m)}} = \mathbf{U}^{(m)} \quad (26)$$

where $\mathbf{W}^{(m)}$ and $\mathbf{b}^{(m)}$ represents the weights and bias in the *m*-th layer. $\mathbf{U}^{(m)}$ can be described as follows:

$$\mathbf{U}^{(m)} = \begin{cases} -\frac{2}{b \times b \times d} \sum_{x=0}^{b-1}\sum_{y=0}^{b-1}\sum_{z=0}^{d-1} \left( X_{x,y,z} - X_{x,y,z}^{(M)} \right) \odot \psi'(\mathbf{a}_{x,y,z}^{(M)}), m = M \\ \left( \begin{array}{l} ((\mathbf{W}^{(M_1+1)})^T \mathbf{U}^{(M_1+1)}) + \\ \lambda_1[2\mathbf{X}^{(M_1)}(\mathbf{I} - \mathbf{Z} - \mathbf{Z}^T - \mathbf{Z}\mathbf{Z}^T)] \end{array} \right) \odot \psi'(\mathbf{a}^{(M_1)}) \\ \qquad + \lambda_2 \sum_{x_i^{(M_1)} \in \mathcal{V}} -\frac{1}{p(y_i / x_i^{(M_1)})} \odot \Phi'(\mathbf{W}_t^{\mathbf{h}1} x_i^{(M_1)} + \mathbf{b}_t^{\mathbf{h}1}) , m = M_1 \\ \left( R_o(\mathbf{W}^{(m+1)} \otimes' \mathbf{U}^{(m+1)}) \right) \odot \psi'(\mathbf{a}^{(m)}), m \in [1, M_1 - l] \cup (M_1 - l, M] \\ \left( (\mathbf{W}^{(m+1)})^T \mathbf{U}^{(m+1)} \right) \odot \psi'(\mathbf{a}^{(m)}), m \in [M_1 - l, M_1 + l] \end{cases}$$

where $\mathbf{a}^{(m)}$ represents the features that are not activated by $\psi$. $R_o$ is the rotation of the input matrix by 180 degree. $\psi'$ is the derivative of $\psi$, and $\Phi'$ is the derivative of $\Phi$. $\odot$ is the element-wise multiplication, and $\otimes'$ is the transposed convolution operation. $\mathbf{I}$ is the identity matrix.

Then, we would perform the update on each iteration:

$$\mathbf{W}^{(m)} = \mathbf{W}^{(m)} - \delta_1 \frac{\partial L_1(\theta)}{\partial \mathbf{W}^{(m)}} \quad (27)$$

$$\mathbf{b}^{(m)} = \mathbf{b}^{(m)} - \delta_1 \frac{\partial L_1(\theta)}{\partial \mathbf{b}^{(m)}} \quad (28)$$

where $\delta_1$ is the learning rate of the 3DCAE.

*2) Optimization for $\mathbf{Z}$:* $\mathbf{Z}$ is computed when $\theta$, $\mathbf{M}$, $\mathbf{h}$ are fixed. As for $\mathbf{Z}$, we can optimize the following problem:

$$\min_{\mathbf{Z}} L_2(\mathbf{Z}) = \min_{\mathbf{Z}} \left\{ \begin{array}{l} \lambda_1 \|\mathbf{X}^{(M_1)} - \mathbf{X}^{(M_1)}\mathbf{Z}\|_F^2 + \lambda_2 \eta \sum_{x_i \in \mathcal{V}} \sum_{x_j \in \mathcal{N}_k(x_i)} \mathbf{h}_2 \mathbf{S}_{i,j} \\ + \frac{\varepsilon}{2} \|\mathbf{Z} - \mathbf{M} + \frac{\mathbf{T}}{\varepsilon}\|_F^2 \end{array} \right\} \quad (29)$$

According to the optimization condition, we have

$$\frac{\partial L_2(\mathbf{Z})}{\partial \mathbf{Z}} = 2\lambda_1 (\mathbf{X}^{(M_1)})^T \mathbf{X}^{(M_1)} (\mathbf{Z} - \mathbf{I})$$

$$+ \lambda_2 \eta \left( \frac{\text{sgn}(\mathbf{Z}+\mathbf{Z}^T)}{2} \mathbf{h}_2^T + \mathbf{h}_2 \frac{\text{sgn}(\mathbf{Z}+\mathbf{Z}^T)}{2} \right) \quad (30)$$

$$+ \varepsilon(\mathbf{Z} - \mathbf{M} + \frac{\mathbf{T}}{\varepsilon})$$

Then, we can obtain

$$\mathbf{Z} = \mathbf{Z} - \delta_2 \frac{\partial L_2(\mathbf{Z})}{\partial \mathbf{Z}} \quad (31)$$

where $\delta_2$ is the learning rate of $\mathbf{Z}$.

*3) Optimization for $\mathbf{M}$, $\mathbf{T}$:* $\mathbf{M}$ can be solved when $\theta$, $\mathbf{Z}$, and $\mathbf{h}$ are fixed. For the $\mathbf{M}$ subproblem, we can optimize the

$$\min_{\mathbf{M}} L_3(\mathbf{M}) = \min_{\mathbf{M}} \beta \|\mathbf{M}\|_1 + \frac{\varepsilon}{2} \|\mathbf{Z} - \mathbf{M} + \frac{\mathbf{T}}{\varepsilon}\|_F^2 \quad (32)$$

Notice that (32) is the $l_1$-$l_2$ term, and it can be directly solved using the shrinkage operator [34]. That is as follows:

$$\mathbf{M}^{(k)}(\bullet, i) = \max \left\{ 0, \|\mathbf{Q}^{(k)}(\bullet, i)\|_2 - \frac{\beta}{\varepsilon} \right\} \frac{\mathbf{Q}^{(k)}(\bullet, i)}{\|\mathbf{Q}^{(k)}(\bullet, i)\|_2} \quad (33)$$

where

$$Q^{(k)} = Z^{(k)} + \frac{T^{(k)}}{\varepsilon}$$

Meanwhile, **T** can be updated by

$$T_{new} = T_{old} + \varepsilon(Z - M) \qquad (34)$$

*4) Optimization for* **h**: For the **h** optimization, with the original objective function that is defined in (17), we can get the following equations:

$$\frac{\partial L_4(\mathbf{h})}{\partial \mathbf{W}^{\mathbf{h}1}} = \sum_{x_i^{(M_1)} \in \mathcal{V}} -\frac{1}{p(y_i/x_i^{(M_1)})} \odot \Phi'\left(\mathbf{W}_t^{\mathbf{h}1} x_i^{(M_1)} + \mathbf{b}_t^{\mathbf{h}1}\right)(x_i^{(M_1)})^T$$

$$\frac{\partial L_4(\mathbf{h})}{\partial \mathbf{b}^{\mathbf{h}1}} = \sum_{x_i^{(M_1)} \in \mathcal{V}} -\frac{1}{p(y_i/x_i^{(M_1)})} \odot \Phi'\left(\mathbf{W}_t^{\mathbf{h}1} x_i^{(M_1)} + \mathbf{b}_t^{\mathbf{h}1}\right)$$
$$(35)$$

and

$$\frac{\partial L_4(\mathbf{h})}{\partial \mathbf{h}_2} = \sum_{x_i^{(M_1)} \in \mathcal{V}} \sum_{x_j^{(M_1)} \in \mathcal{N}_k\left(x_i^{(M_1)}\right)} -S_{i,j} \qquad (36)$$

Therefore, we can update $\mathbf{h}_1$ and $\mathbf{h}_2$ as follows:

$$\mathbf{h}_1 = \mathbf{h}_1 - \tau \frac{\partial L_4(\mathbf{h})}{\partial \mathbf{h}_1}, \quad \mathbf{h}_2 = \mathbf{h}_2 - \tau \frac{\partial L_4(\mathbf{h})}{\partial \mathbf{h}_2} \qquad (37)$$

where $\tau$ is the step size.